B. Comp Dissertation

# Explainable Fraud Detection with GNNExplainer and Shapley Values

By

HIEU, DAO NGOC

Department of Computer

ScienceSchool of Computing

National University of

Singapore2022/2023

B. Comp Dissertation

# Explainable Fraud Detection with GNNExplainer and Shapley Values

By

HIEU, DAO NGOC

# Abstract

The risk of financial fraud is increasing as digital payments are used more and more frequently. Although the use of artificial intelligence systems for fraud detection is widespread, society and regulators have raised the standards for these systems' transparency for reliability verification purposes. To increase their effectiveness in conducting fraud investigations, fraud analysts also profit from having concise and understandable explanations. To solve these challenges, the paper will concentrate on developing an explainable fraud detector.

Subject Descriptors:

- I.2     Artificial Intelligence
- H.5.2   User Interfaces
- I.5.1   Models
- I.5.2   Design Methodology
- J.1     Administrative Data Processing

Implementation Software and Hardware:
   Python, PyTorch 2.0,
   TorchGeometric 1.13, Bash Script

# Acknowledgement

I want to express my gratitude to my supervisor, Brian Lim Youliang, for his constant encouragement and advice throughout this semester. I would also want to thank my project partner, Xuan Zhi, for the fruitful and motivating conversations and ongoing assistance with the project.

# Table of Contents

# 1. Introduction

## 1.1. Motivation

Our spending habits are changing as a result of the emergences of digital payment systems, and the security threats that come with it are not insignificant. Account takeover, financial information theft, fraudulent chargebacks, money laundering, and many more issues are typical dangers existing in the digital space. These illegal actions can substantially harm the platform's credibility, have a detrimental influence on user experiences, and result in financial losses. As a result, it is crucial to spot fraudulent practices and take all reasonable precautions to reduce risks.

Fraud detection and decision support systems are widely used to eliminate fraud from the ecosystem of digital banking and e-commerce. A significant percentage of these systems employ Artificial Intelligence (AI) to better recognize subtle and complex patterns of fraud based on historical and real-time data. However, users of such systems are presented with a black-box problem, and thus, they might deem predictions from AI untrustworthy. Existing solutions for this issue include Explainable AI (XAI) which provides experts with explained predictions through various methods, such as subgraph and feature importance explanation. Nevertheless, implementing XAI has its own set of challenges. XAI requires semantic financial data which is often inaccessible due to privacy and security concerns. Additionally, explanations generated by most existing XAI are usually vague, unintuitive, and open to potentially contradicting interpretations.

## 1.2. Project Objective

To overcome the aforementioned challenges, this project aims to investigate and propose a machine learning model for fraud detection accompanied by an explanation method. The target data is the tabular transaction data containing typical information such as sender, receiver, monetary amount and timestamp. The explanation is designed to simulate an actual fraud investigation, presenting all relevant evidence and patterns for a data analyst to come up with a plain-text explanation with little effort. The project mainly focus on the interpretability of the fraud detector.

The main objectives:
1. Propose a machine learning model to detect consumer payment frauds

2. Provide a clear explanation for the prediction of the model by utilizing finance knowledge and data visualization

# 2. Background

## 2.1. Consumer Payment Fraud

Consumer payment fraud falls into one of two basic categories: authorized or unauthorised. Unauthorized transactions are more prevalent in volume, at least in the UK, where more than £360 million in thefts were reported in H1 2022 alone, according to UK Finance (2021). Unauthorized fraud occurs when the account holder does not give consent and a criminal completes the transaction, for example, by using the victim's card information without their knowledge or approval which is usually referred to as account takeover fraud (ATO). This project will focus on detecting and explaining ATO frauds.

According to Forbes Finance Council (2022), ATO is the process by which criminals use a number of techniques, such as buying stolen information from the dark web, social engineering, phishing, password cracking, or credential stuffing, to take control of online accounts that do not belong to them in order to further a variety of malicious objectives. Unfortunately, this attack strategy has seen an uptick recently. ATO fraud now ranks as the fifth most frequent attack method for North American merchants after occurring in the past year for 27% of worldwide merchants who took part in the 2022 Global Payments and Fraud Report (Merchant Risk Council, 2022).

## 2.2. Machine Learning and Explainable AI in Fraud Detection

For digital payment service providers, fraud detection has been a crucial topic. It is researched in several applications, including anti-money laundering (Emmerich, Pudelko, Gallenmüller, and Carle, 2017), spam reviews and news identification (Shu, Mahudeswaran, Wang, and Liu, 2020), fraudulent account detection e.g. social networks (Breuer, Eilat, and Weinsberg, 2020), online payment systems (Zhong et al, 2020), and online store platforms (Rao, 2020). While most of the research proposed machine learning models with high accuracy and scalability, their interpretability appears to be a significant concern. Detecting a fraudulent transaction is not a trivial process, and false predictions could negatively affect customer experience and the service provider's credibility. Typically, this process requires costly and inefficient human investigation to review the model's prediction. Therefore, automatically generated explanation is necessary to eliminate human intervention in fraud detection.

Homogeneous graph has been widely applied in fraud prevention systems e.g. anti-money laundry (Weber et al, 2019), risky/malicious account detection (Liang et al, 2019). Solutions to real-world anomaly detection problems using heterogeneous graph (Liu et al, 2018) or to combine homogeneous and heterogeneous graphs (Li, Qin, Liu, Yang, and Li, 2019) have become more prevalent, because it allows aggregating information propagation through different node and edge types. GEM by Liu et al (2018) employed attention mechanism in a device-account heterogeneous graph to capture user activities and device embeddings in each subgraph neighbourhood, and the model performs very well for Alipay in its day-to-day business, despite the fact that it does not consider type features as inputs.

In Rao (2020), a Graph Neural Network (GNN) (Gori, Monfardini, and Scarselli, 2005) model is used to detect fraudulent transactions on eBay - an online marketplace. Subsequently, GNNExplainer (Ying, Bourgeois, You, Zitnik, and Leskovec, 2019) is utilized to generate a subgraph around the node-to-explain as an explanation for the GNN model's prediction. This is a good visualization aid for a data analyst to arrive at the final plain-text explanation, considering that one of the biggest finance company in the world also offers graphs as its main tool for fraud investigation (SAS Institute Inc., n.d). However, in most cases, having just a static graph makes the explanation vague and open to potentially conflicting interpretation. To alleviate this problem, additional explanation methods could be added together with GNNExplainer to provide more evidence, such as SHAP (Lundberg and Lee, 2017).

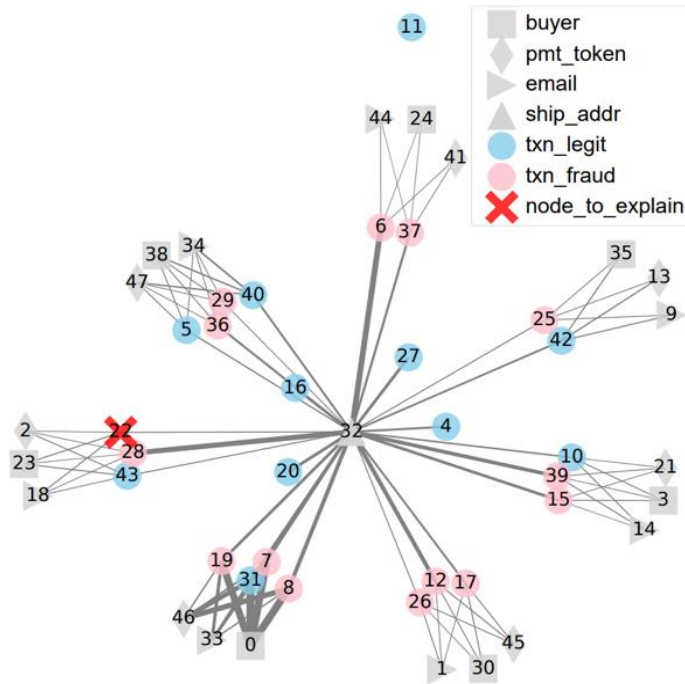

Figure 1: xFraud graph-based explanation

# 3. Literature Review

The development of interpretable machine learning tasks for use in fraud detection is still in progress, and more academics are working to maximize their utility. In this section, the research is discussed in two parts, the machine learning methods for fraud detection and the interpretable methods used in these models. The fraud detection methods are covered in Section 3.1. The approaches utilized to ensure the interpretability of the machine learning model are covered in Section 3.2.

## 3.1. Machine Learning Models for Fraud Detection

### 3.1.1. Transformer

Existing machine learning models relying on CNNs as basic building blocks require computation of representations for all input and output positions. However, in these models, the number of operations involved in the computation grows as the distance between positions increases, making it computationally expensive to quantify dependencies between distant entities in a network. Transformer (Vaswani et al, 2017) is proposed as a solution to this problem by introducing a method to compute position representations with a constant number of operations.

The concept of an attention function is described as mapping a query and a set of key-value pairs to an output. The output is a weighted sum of the values whose weights are computed by quantifying how compatible the query is with the corresponding key. In short, the attention mechanism looks at an input sequence and decides at each step which other parts of the sequence are important.

Multiheaded attention is introduced to allow the model to control a combination of representations from different positions at the same time as well as to increase performance. In a multiheaded attention module, each single attention calculation is carried out in parallel with different projections of key, value, and query vectors. The resulted vector from each head is concatenated and projected to the original dimension of the input vectors. Although Transformer is primarily used in natural language processing applications, we could adapt this in our GNN to represent relationships among nodes as attention vectors.

### 3.1.2. Heterogeneous Graph Transformer

The difficulties associated with heterogeneous graph processing in existing GNNs are addressed by the introduction of HGT (Hu, Dong, Wang, and Sun, 2020). HGT leverages heterogeneous mutual attention by analysing each edge $e = (s, t)$ based on its meta relation triplet, which consists of the node type of s, the edge type of e between s and t, and the node type of t. This method avoids giving parameters to specific types of edges. This method enables different types of nodes and edges to keep their distinctive representation spaces. Furthermore, the HGT architecture enables nodes of various types to interact, transmit, and aggregate messages without being bound by the discrepancies in their distributions. HGT may integrate data from high-order neighbours of various types by message passing across layers, which can be thought of as "soft" meta routes. HGT's attention mechanism may automatically and implicitly learn and identify "meta pathways" that are important for various downstream tasks even if it is just given one-hop edges as input without explicitly constructing meta paths.

There are 3 main components of HGT 1) Heterogeneous Mutual Attention 2) Heterogeneous Message Passing and 3) Target-Specific Aggregation.

In the first component, HGT computes the mutual attention between source and target node of an edge by aggregating weighted messages sent by neighboring nodes. The weights are attention vectors whose computation is inspired by Transformer. The second component is carried out in parallel with the first one. For a pair of nodes, a single message is calculated by multiplying the projected representation of the source node with the edge dependency matrix, which will then be concatenated to get the final message for each node pair. The last component aggregates the attention and message vectors obtained in the previous 2 component to output the final representation of a target node from the information of its neighboring nodes.

HGT uses different weight matrices for different relations, thus it is better than the original Transformer in handling distribution differences in heterogeneous graphs.

### 3.1.3. xFraud Detector

The xFraud detector model is inspired by Transformer (Vaswani et al, 2017) and HGT (Hu et al, 2020). xFraud detector uses the self-attention mechanism as presented in Transformer but adapts it for GNN as shown in HGT. Compared to HGT, xFraud detector does not use different weight matrices for different node and edge types to improve performance while model accuracy is maintained. Moreover, xFraud does not adopt temporal encoding in HGT to allow the model to view all of transactions made by a user at any point in time, as well as other relevant entities linked to the transactions.

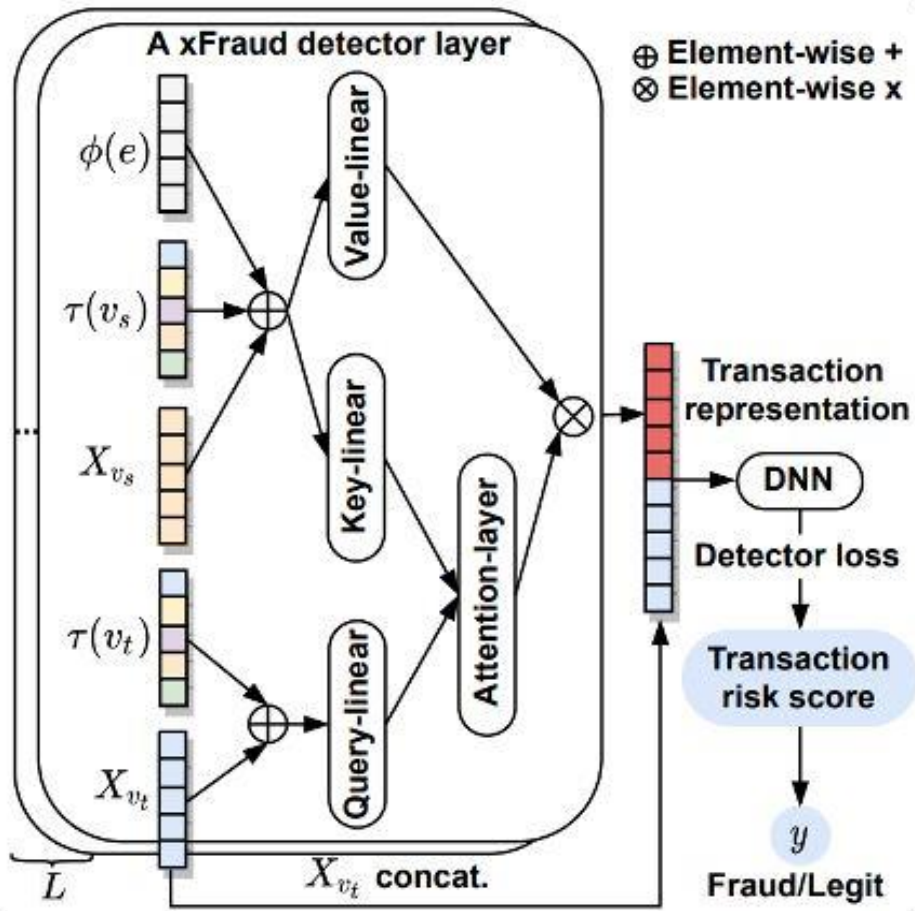

Figure 2: xFraud detector

The xFraud has 4 steps: 1) The detector received a heterogeneous graph including node features, one-hot encoded node types and edge types as inputs. For non-transaction nodes, the initial feature values are 0; 2) Convolution layers with self-attention mechanism (Vaswani et al, 2017): each layer obtains a set of query, key and value vectors from the above inputs. Attention scores are calculated for the source and target node of each edge, then layer-wise normalized. These scores are fed into a *ReLU* activation function which are taken by the next layer as inputs; 3) A *tanh* activation is applied to the final transaction representation after the above convolution layers whose outputs are concatenated with the original transaction features and fed into a feedforward connected network with 2 hidden layers. Dropout, layer normalization, and *ReLU* transformation are applied to get a risk score; 4) The loss function used is the cross entropy of the fraud label and the probability score calculated by *softmax*.

## 3.2. Explainable Machine Learning Methods

### 3.2.1. LIME

LIME (Ribeiro, Singh, and Guestrin, 2016) stands for Local Interpretable Model-agnostic Explanations. This algorithm samples instances close to the original instance, and trains a simpler model (e.g. a linear model) with these samples. The results from training the smaller model are then examined to craft explanations for the original model. Nonetheless, LIME faces many challenges including difficulties in sampling data points and choosing the right explanation model. LIME is model-agnostic and easy to use, so it can be used as a baseline for more complex explanation methods.

### 3.2.2. Shapley Values

Shapley Values (Shapley, 1953) is a method from the coalitional game theory which provides the contribution of each feature of an instance to its prediction. More specifically, the Shapley value of a feature is its average marginal contribution across all possible combinations. To calculate the Shapley value for a feature $v$, predictions are calculated with and without $v$ together with all combinations of the values of other features and take the difference to get the marginal contribution.

Shapley Values is better than LIME (Ribeiro et al, 2016) regarding the fair distribution among the feature values of an instance. However, this method requires a lot of computing time ($\Omega(2^n)$), and in a typical dataset with hundreds of features, it is not feasible to calculate Shapley Values.

### 3.2.3. SHAP and KernelSHAP

SHAP (Lundberg et al, 2017) stands for Shapley Additive Explanations which was proposed by Lundberg and Lee. SHAP aims to approximate Shapley Values (Shapley, 1953) since the original method is too inefficient in real-world settings. Lundberg and Lee (2017) proposed several approaches including KernelSHAP. KernelSHAP generates samples based on feature masks where present features are mapped to the instance's features, while absent features are taken from a random data instance. KernelSHAP's weight assignment is different from LIME (Ribeiro et al, 2016) which weights instances according to their distance to the original instance; instead, KernelSHAP weights feature masks based on the missingness/isolation of features in the mask, as the algorithm learns more when features are isolated.

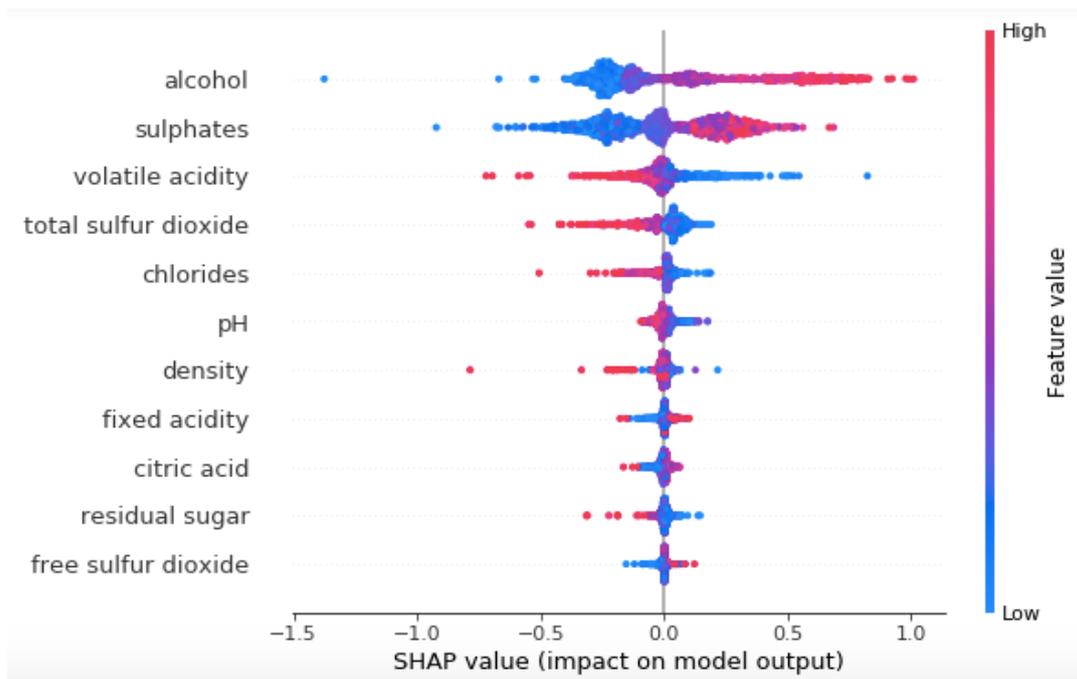

Figure 3. A sample SHAP values plot

For example, Figure 1 is made of all the dots in the train data. It delivers the following information:

- *Feature importance:* features are ranked in descending order.
- *Impact:* The horizontal location shows whether the effect of that value *is associated with a higher or lower prediction*.
- *Original value:* Colour shows whether that variable is high (in red) or low (in blue) for that observation.
- *Correlation:* A *high* level of the "alcohol" content has a high and *positive* impact on the quality rating. The "high" comes from the red colour, and the "positive" impact is shown on the X-axis. Similarly, we will say the "volatile acidity" is negatively correlated with the target variable.

3.2.4. GNNExplainer

GNNs are especially useful for machine learning on graph datasets. The key feature of GNNs is pairwise message passing where graph nodes sequentially update their features by exchanging information with neighbouring nodes, capturing graph-level data in the process. However, GNNs does not provide sufficient transparency in their predictions due to the typically large number of layers in models leveraging GNNs. GNNExplainer (Ying et al, 2019) is a solution for this problem where a trained GNN and its predictions are taken as inputs and the output is a subgraph with a subset of features that are most important in predicting the results. GNNExplainer is model-agnostic and can be used to provide more insights into predictions from any GNN model.

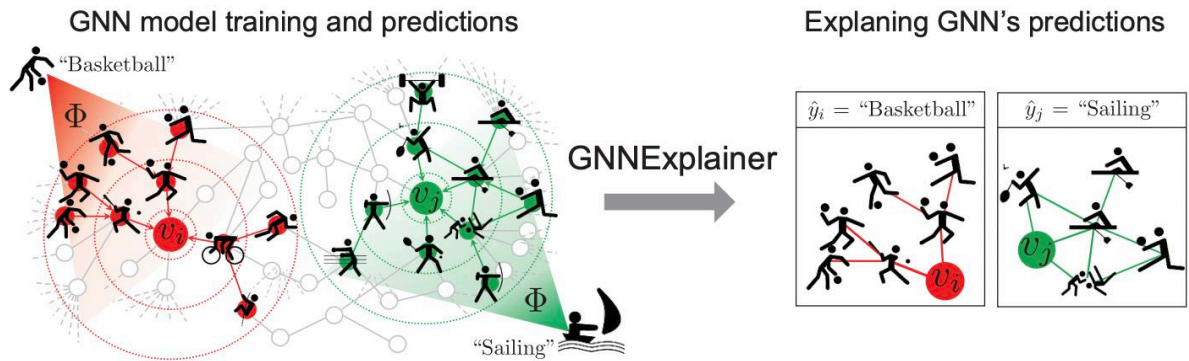

Figure 4. Subgraphs generated for 2 nodes $v_i$ and $v_j$, and nodes in each subgraph correspondingly support the predictions of $v_i$ and $v_j$

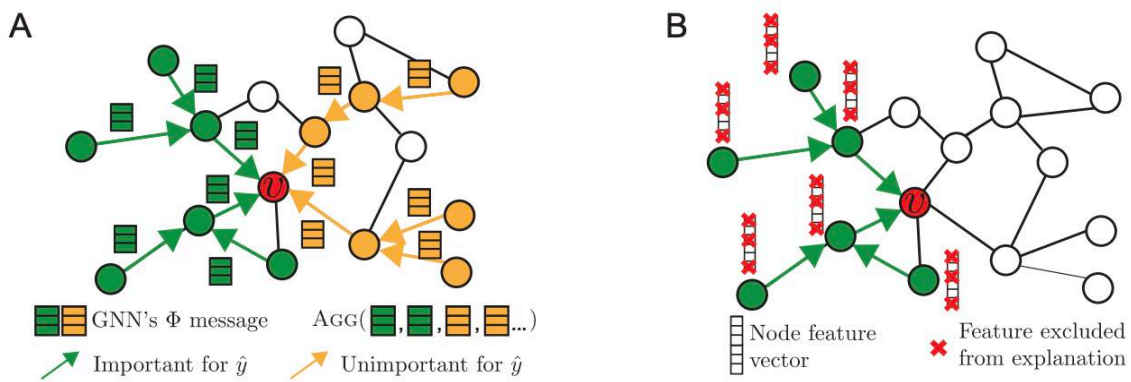

Figure 5. Based on the importance of neural message-passing pathways (green represents important paths, yellow represents unimportant ones), GNNExplainer generates the set of most influential features for node $v$'s prediction

GNNExplainer generates a subgraph of the entire original input graph such that the shared information between the two graphs is maximized. For each node in the original graph, mutual information (MI) is calculated which quantifies the change in the final prediction of the node when the node's computation graph is limited to the current explanation subgraph. The process of generating a subgraph $G_S$ with a subset of associated features $X_S$ is formulated as maximizing

MI between the subgraph with associated features and the original graph G and whole set of features X:

$$\max_{G_S} MI\left(Y, (G_S, X_S)\right) = H(Y) - H(Y|G = G_S, X = X_S)$$

where $H$ is the entropy function. At the end of the process, an edge mask and a feature mask will be generated to indicate the edges and features significant to the prediction respectively and their weights. However, using MI makes it difficult to understand the direction of an edge's or feature's effect on a class prediction job, as MI scores can only be positive. As such, it is unclear if a certain node feature or edge has negative or positive correlation with the prediction of a node.

## 4. Research Method

### 4.1. Datasets

We obtain the synthetic payment data for fraud detection from J.P.Morgan. The dataset is generated based on the multi-agent framework proposed by Borrajo et al. (2020). The simulator used by planning agents is made up of different parts, including Execution. This module works by taking in information about the environment and using it to generate a plan through a reasoning cycle that involves Planning, executing actions, and observing the resulting state. Goal generation is also used to obtain new goals or state components. The simulator can be used with different domains, except for Goal reasoning which requires generating behaviour for at least two types of agents within the same domain.

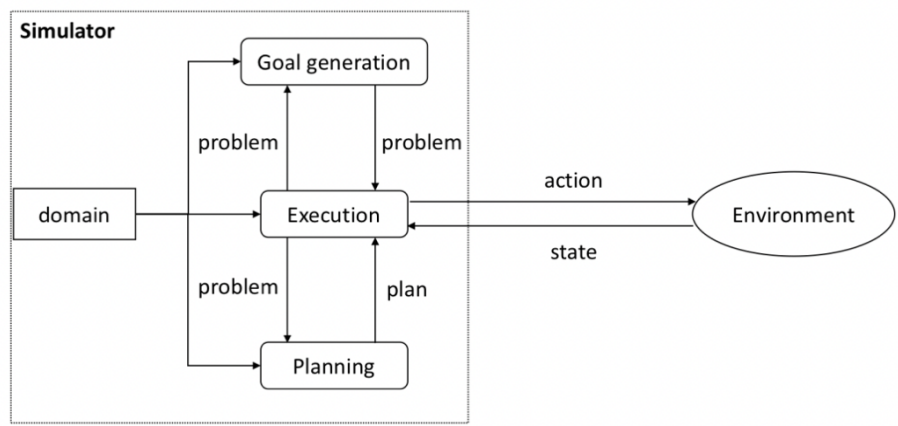

Figure 6: High-level view of the simulator

However, the framework above is domain-independent, and the exact details of behaviour patterns of agents in the dataset are not provided by J.P.Morgan, so further data analysis is necessary to identify fraudulent patterns.

| Transaction_Id | Sender_Id | Sender_Account | Sender_Country | Sender_Sector | Sender_lob | Bene_Id | Bene_Account | Bene_Country | USD_Amount | label | Transaction_Type |
|---|---|---|---|---|---|---|---|---|---|---|---|
| PAY-BILL-3589 | CLIENT-3566 | ACCOUNT-3578 | USA | 21264 | CCB | COMPANY-3574 | ACCOUNT-3587 | GERMANY | 492.67 | 0 | MAKE-PAYMENT |
| WITHDRAWAL-3591 | CLIENT-3566 | ACCOUNT-3579 | USA | 18885 | CCB | | | | 388.92 | 0 | WITHDRAWAL |
| MOVE-FUNDS-3528 | CLIENT-3508 | ACCOUNT-3520 | USA | 4809 | CCB | COMPANY-3516 | ACCOUNT-3527 | GERMANY | 280.7 | 0 | MOVE-FUNDS |
| WITHDRAWAL-3529 | CLIENT-3508 | ACCOUNT-3519 | USA | 7455 | CCB | | | | 118.14 | 0 | WITHDRAWAL |
| QUICK-DEPOSIT-3471 | | | | | | CLIENT-3442 | ACCOUNT-3461 | USA | 105.16 | 0 | DEPOSIT-CASH |
| QUICK-DEPOSIT-3473 | | | | | | CLIENT-3442 | ACCOUNT-3460 | USA | 164.97 | 0 | DEPOSIT-CASH |
| PAY-BILL-3404 | CLIENT-3384 | ACCOUNT-3395 | USA | 36316 | CCB | COMPANY-3392 | ACCOUNT-3401 | GERMANY | 456.89 | 0 | MAKE-PAYMENT |
| QUICK-DEPOSIT-3406 | | | | | | CLIENT-3384 | ACCOUNT-3396 | USA | 413.17 | 0 | DEPOSIT-CASH |
| PAY-CHECK-3347 | CLIENT-3330 | ACCOUNT-3341 | USA | 36194 | CCB | CLIENT-3333 | ACCOUNT-3338 | CANADA | 377.65 | 0 | PAY-CHECK |
| PAY-CHECK-3348 | CLIENT-3330 | ACCOUNT-3340 | USA | 20626 | CCB | CLIENT-3333 | ACCOUNT-3338 | CANADA | 338.03 | 0 | PAY-CHECK |
| MOVE-FUNDS-3292 | CLIENT-3272 | ACCOUNT-3284 | USA | 21568 | CCB | CLIENT-3275 | ACCOUNT-3291 | CANADA | 100.85 | 0 | MOVE-FUNDS |
| MOVE-FUNDS-3294 | CLIENT-3272 | ACCOUNT-3284 | USA | 29040 | CCB | CLIENT-3273 | ACCOUNT-3289 | USA | 276.66 | 0 | MOVE-FUNDS |
| PAY-BILL-3232 | CLIENT-3203 | ACCOUNT-3222 | USA | 27393 | CCB | COMPANY-3210 | ACCOUNT-3218 | GERMANY | 234.88 | 0 | MAKE-PAYMENT |
| QUICK-DEPOSIT-3234 | | | | | | CLIENT-3203 | ACCOUNT-3222 | USA | 945.22 | 0 | DEPOSIT-CASH |
| DEPOSIT-CASH-3163 | | | | | | CLIENT-3139 | ACCOUNT-3154 | USA | 655.09 | 0 | DEPOSIT-CASH |
| PAY-BILL-3162 | CLIENT-3139 | ACCOUNT-3153 | USA | 25066 | CCB | COMPANY-3147 | ACCOUNT-3160 | GERMANY | 675.37 | 0 | MAKE-PAYMENT |
| WITHDRAWAL-3100 | CLIENT-3075 | ACCOUNT-3090 | USA | 22778 | CCB | | | | 319.95 | 0 | EXCHANGE |
| QUICK-PAYMENT-3099 | CLIENT-3075 | ACCOUNT-3091 | USA | 39013 | CCB | CLIENT-3078 | ACCOUNT-3087 | TAIWAN | 771.54 | 0 | QUICK-PAYMENT |
| PAY-BILL-3036 | CLIENT-3016 | ACCOUNT-3028 | USA | 43951 | CCB | COMPANY-3022 | ACCOUNT-3033 | GERMANY | 730.69 | 0 | MAKE-PAYMENT |

Figure 7: Sample data from the synthetic J.P.Morgan dataset

## 4.2. Data Analysis and Feature Engineering

Data analysis is carried out using JMP Pro (Jones and Sall, 2011) software application. We suspect that the differences in transaction amount and timestamp between different transactions of the same sender or receiver could be good features to the machine learning model. We label them *dAmount* and *dTime* accordingly. Non-fraud transactions are labelled 0 and fraud transactions are labelled 1.

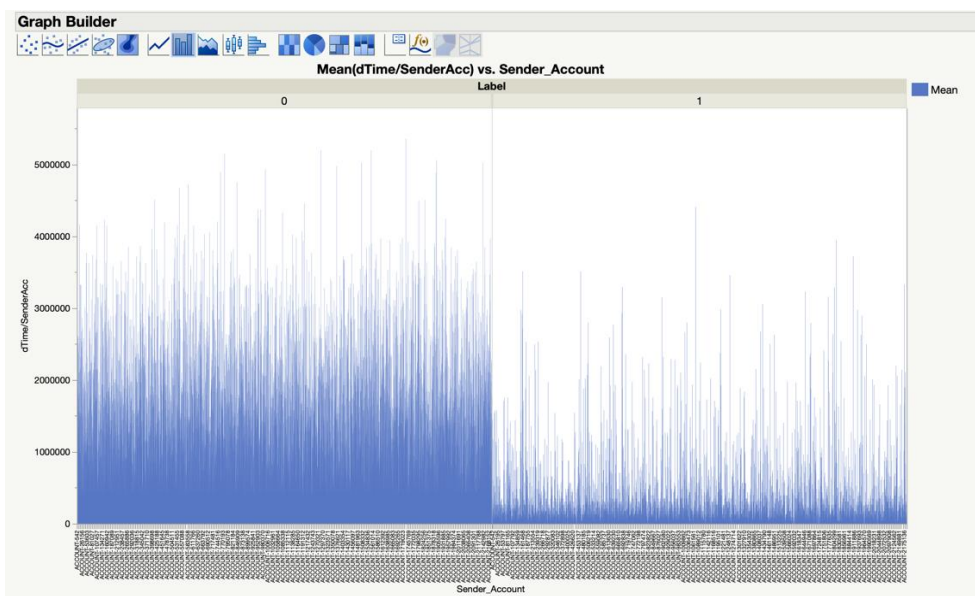

Figure 8: Bar chart for *dTime* against fraud and non-fraud accounts

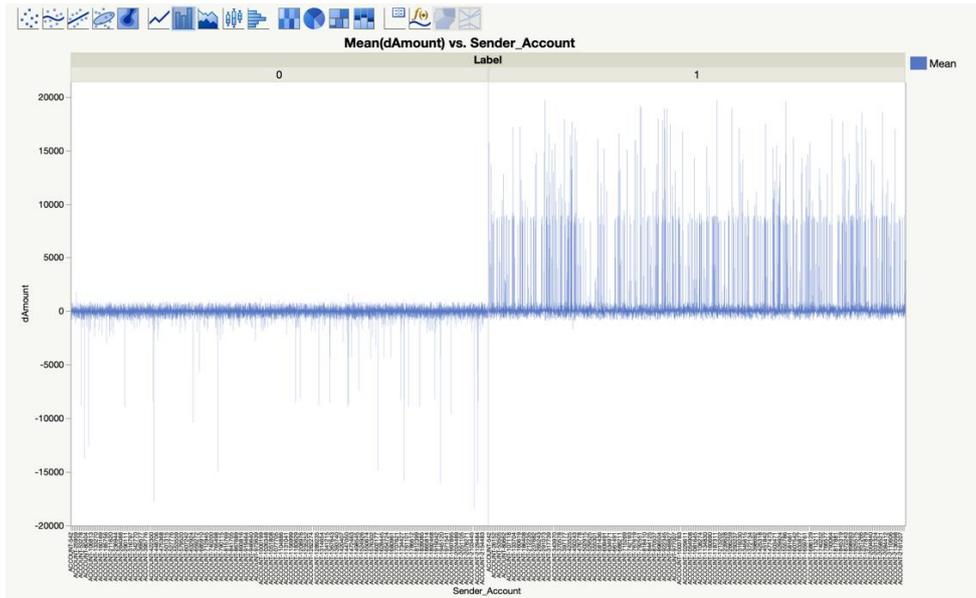

Figure 9: Bar chart for *dAmount* against fraud and non-fraud accounts

From the plots above, there are clear differences in these 2 features between the 2 classes. Non-fraud accounts seem to take more time between transactions, while fraud accounts transact quite rapidly. The differences between transaction amounts of fraud accounts are quite significant, while those of non-fraud accounts are mostly stable. We decide to add these 2 features to the original dataset.

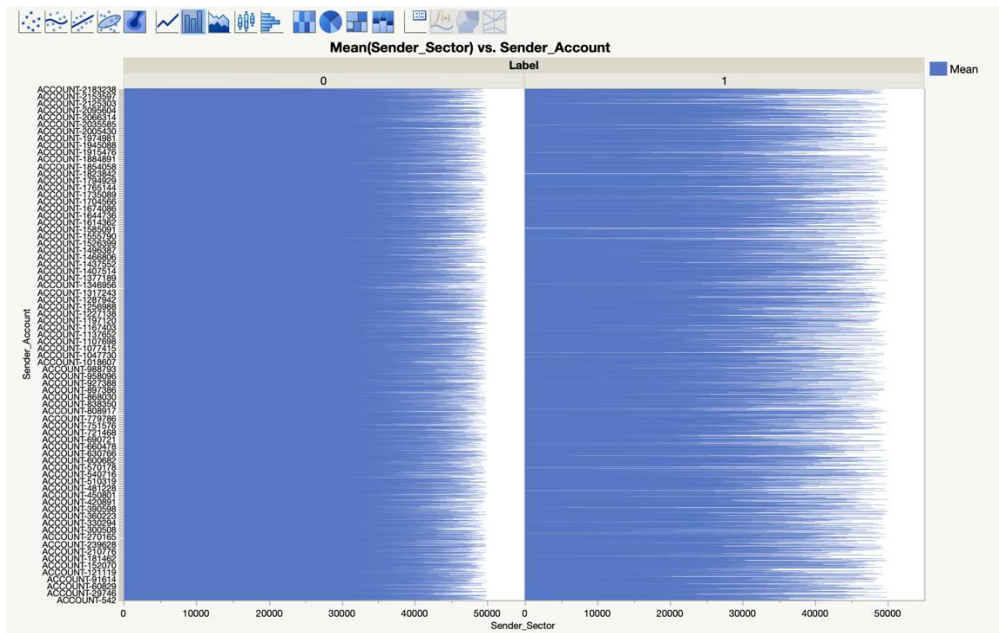

Figure 10: Distribution of values of Sector field for fraud and non-fraud accounts

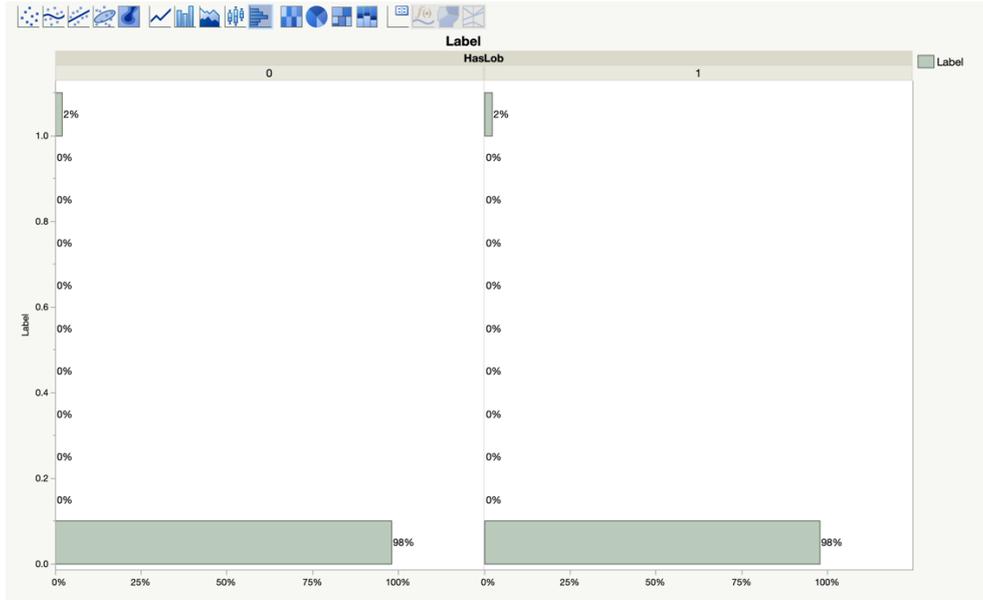

Figure 11: Distribution of values of Line of Business field for fraud and non-fraud accounts

The *Sector* and *Lob* (Line of Business) fields contain abstract values which might not aid in explaining the predictions. Additionally, as Figure 10 and 11 illustrate, there is not much difference of the value distribution of these field between fraud and non-fraud cases, so we remove them from the dataset.

4.3. Graph Construction

In a heterogeneous graph $G$, each node $v \in V$ has a type $NodeType(v) \in A$ where $A \coloneqq \{Transaction, Account, User, Country\}$. For $Transaction$ nodes, their initial features are $Features(v) \coloneqq \{USD\ Amount, dAmount, dTime\}$. Node and edge type features are one-hot encoded before being fed into the heterogeneous convolution layer.

We extract the nodes from each row. For example, an *Account* node is created for each unique value of *Sender_Account* and *Bene_Account* in the dataset, and a *Transaction* node is created according to the *Transaction_Id* field. *Transaction* nodes are mapped to non-transaction nodes extracted from the same rows, resulting in edges starting at the transaction node and ending at corresponding non-transaction nodes.

Regarding edge types, we have the following mappings between edge types and involved node types:

$$Executed\_In: Transaction - Country$$
$$Sent\_To, Sent\_By: Transaction - Account$$
$$Transferred\_By, Received\_By: Transaction - Users$$

We construct the graph based on the tabular dataset from J.P.Morgan using Neo4j database.

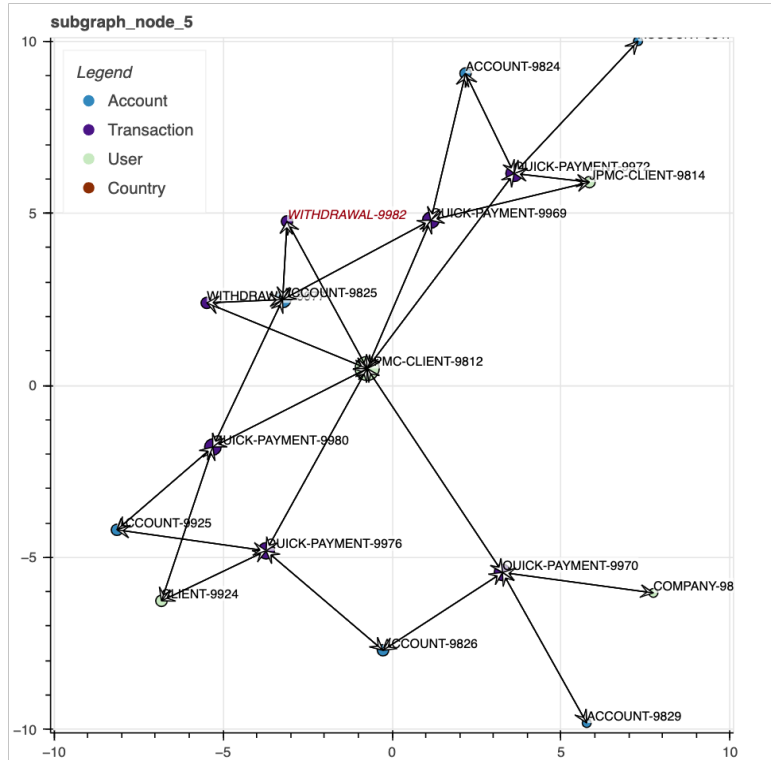

Figure 12: A sample subgraph of the constructed graph dataset

### 4.4. Baseline Models

GEM (Liu et al, 2018) and GAT (Veličković et al, 2017) are chosen as baseline models due to them being similar to the proposed model while having simpler architecture.

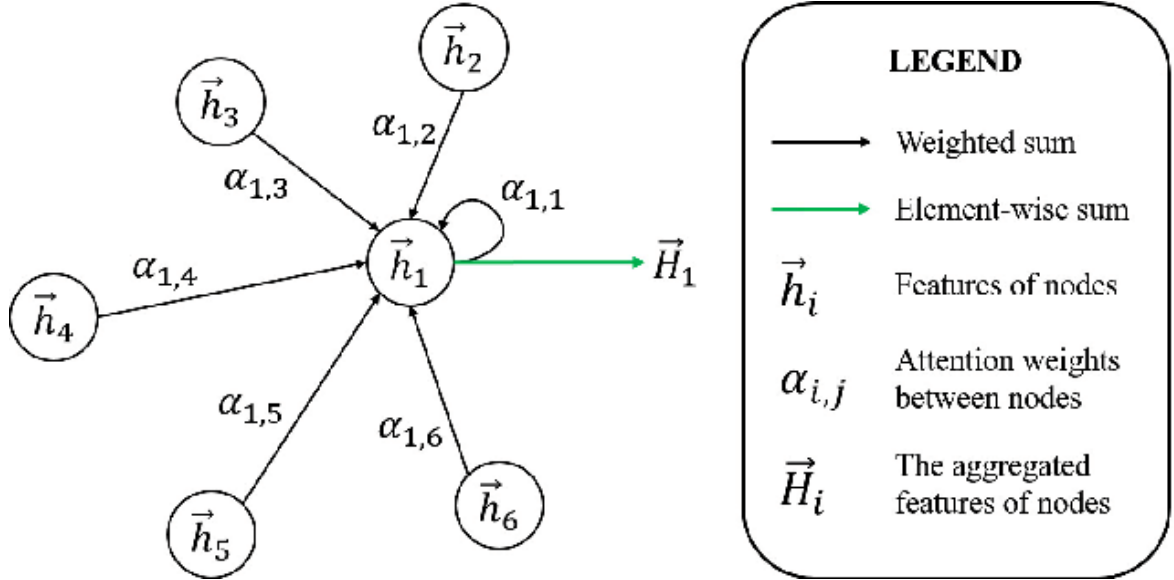

Figure 13: Graph Attention Network

Similar to the proposed model, GAT uses attention mechanism to compute node representation in a graph. However, GAT does not take into account node type and edge type features in its attention calculation, so it does not perform as well as the proposed model facing heterogenous graph inputs. Moreover, GAT calculate attention for each node sequentially, whereas the proposed model adopts the multiheaded attention mechanism from Transformer (Vaswani et al, 2017). Nevertheless, GAT could still be a good baseline model for the task of fraud detection.

Regarding GEM, while it targets heterogeneous graph inputs, its node embedding does not contain type features, and thus it cannot learn interaction between type relations. Fortunately, this model has been experimented on Alipay real-world datasets, and used in Alipay's payment system to detect fraud, so GEM is a good baseline model for our propose model.

### 4.5. Proposed Model

We utilize the xFraud detector model presented in section 3.1.3 with some modifications to its heterogeneous convolution layer.

For a tuple $[NodeType(v_s), EdgeType(e), NodeType(v_t)]$, where $e = (v_s, v_t)$, we initialize the node type embeddings $NodeTypeEmb(v)$ and edge type embeddings $EdgeTypeEmb(e)$ with zero weights as well as the attention weight matrices $W^{att}_{NodeType(v_s)}$ and $W^{att}_{NodeType(v_s)}$ for source and target node respectively. The weight matrices for key, query, and value vectors denoted by $W^K, W^Q, W^V$ are initialized using the Xavier method (Glorot and Bengio, 2010).

A heterogeneous convolution layer of the target node $v_t$ computes its representation as follow:

$$H^l[v_t] \leftarrow Aggregate(Attention(v_s, v_t) \cdot Message(v_s))$$

$$Aggregate(*) = \sigma(Mean(*))$$

where $H^l[v_t]$ is the representation of target node $v_t$ on layer $l$. The *Aggregate* function carries out a simple averaging followed by a non-linear activation to aggregate neighbourhood message by the attention weight.

For each target node $v_t$, we create *query, key* and *value* vectors for self-attention mechanism with multiheads.

To construct the *query* vector for the target node $v_t$ in the $i^{th}$ head, we start with inputs to the first layer by taking transaction features $Features(v_t)$ of the target node and its node type embedding $NodeTypeEmb(v_t)$ to compute the following:

$$Q^i(v_t) = QLinear^i_{NodeType(v_t)}(Features(v_t) + NodeTypeEmb(v_t))$$

For subsequent layers, we use the results from the previous layer to compute the *query* vector as follow:

$$Q^i(v_t) = QLinear^i_{NodeType(v_t)}(H^{l-1}[v_t])$$

where $H^{l-1}[v_t]$ is the representation of target node on the $l-1$ layer.

Similarly, for the *key* and *value* vectors, on the first heterogeneous convolution layer, we do the following:

$$K^i(v_s) = KLinear^i_{NodeType(v_s)}(Features(v_s) + NodeTypeEmb(v_s) + EdgeTypeEmb(e))$$

$$V^i(v_s) = VLinear^i_{NodeType(v_s)}(Features(v_s) + NodeTypeEmb(v_s) + EdgeTypeEmb(e))$$

For subsequent layers, we use the previous layers' results as follow:

$$K^i(v_s) = KLinear^i_{NodeType(v_s)}(H^{l-1}[v_t])$$

$$V^i(v_s) = VLinear^i_{NodeType(v_s)}(H^{l-1}[v_t])$$

After obtaining the *query, key* and *value* vectors, we compute the attention head:

$$AttentionHead^i(v_s, v_t) = \frac{(K^i(v_s)W^{att}_{NodeType(v_s)} + Q^i(v_t)W^{att}_{NodeType(v_t)})}{\sqrt{d_k}}$$

where $d_k$ is the length of the *key* vector.

Then we can obtain the final attention vector by combining all the attention heads above, and apply the *softmax* activation function over the target node's neighbouring nodes:

$$Attention(v_s, v_t) = \underset{\forall v_s \in N(v_t)}{softmax}(\underset{i \in [1,h]}{\|} AttentionHead^i(v_s, v_t))$$

Where $N(v_t)$ is the neighbours of the target node, $h$ is the number of attention heads, and $\|$ is vector concatenation operator.

Finally, messaging passing between 2 consecutive layers is computed as follow:

$$Message(v_s) = \underset{i \in [1,h]}{\|} (V^i(v_s) \cdot dropout(AttentionHead^i(v_s, v_t)))$$

Overall, the xFraud detector is very similar to HGT (Hu et al, 2020), but some differences in attention and message passing computation are made as we do not want to carry out aggregation on different node types by using the different weight matrices. This gives us better performance in the detector on our synthetic dataset.

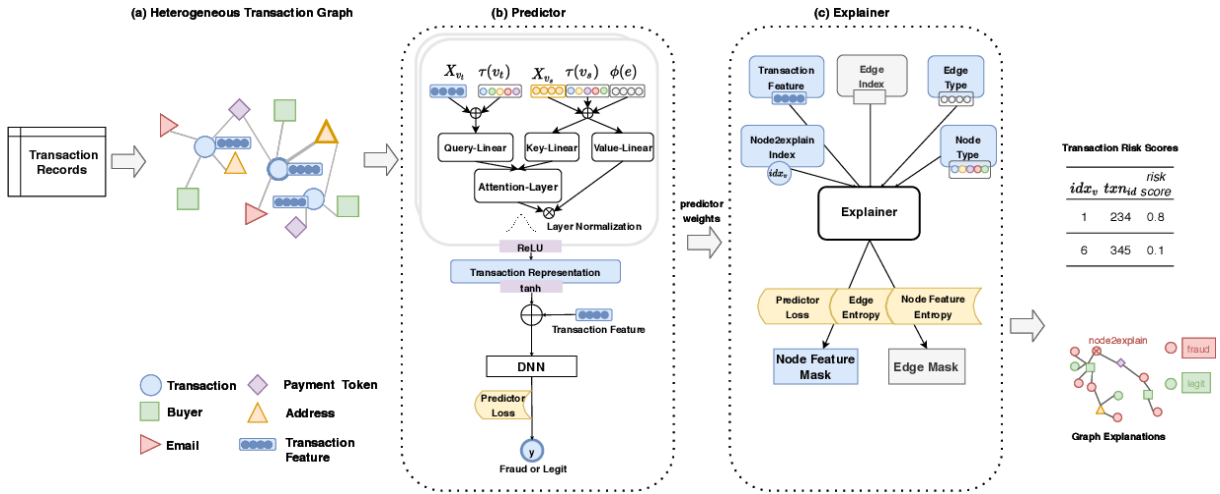

Figure 14: The complete end-to-end architecture of the proposed model

## 4.5. Composite Explainer

We propose a composite explainer consisting of 3 components 1) a GNNExplainer graph, 2) node feature importance, and 3) edge missingness importance.

### 4.5.1. GNNExplainer

To provide explanations for the predictions made by our fraud detection model, we utilized GNNExplainer (Ying et al, 2019). GNNExplainer takes in the trained xFraud detector model, the node to explain $v$ and the input subgraph relevant to the node $v$, and produces edge and feature weights relative to the node to explain. These weights are used to extract the most important edges and nodes to be included in the final explanation subgraph. To further ensure a concise explanation subgraph, the maximum depth for a subgraph is 6, and the maximum width at each level is 16, making the maximum number of nodes in a graph 96. This is consistent with the input to the detector for each node.

As previously discussed in Section 3.2.4, it is challenging to determine the directional influence of selected nodes and features on the model's prediction using MI. Thus, we proposed an algorithm to approximate Shapley values to quantify edge missingness and node feature contribution to the prediction. To improve the performance of this proposed algorithm, we implemented an approximation algorithm proposed by Strumbelj and Kononenko (2014) with Monte-Carlo Sampling:

$$\widehat{\phi_j} = \frac{1}{M} \sum_{m=1}^{M} \left( \hat{f}(x_{+j}^m) - \hat{f}(x_{-j}^m) \right)$$

### 4.5.2. Node Feature Shapley Values

The algorithm to approximate Shapley values for single node feature value is as follows:
- Output: Shapley value for the value of the $j^{th}$ feature
- Required: Number of iterations M, node of interest x, feature index j, subgraph G, and machine learning model f
    - For all m = 1,…,M:
        - Draw random node z from the subgraph G
        - Choose a random permutation o of the feature values

- Order node x's features: $x_o = (x_{(1)}, \ldots, x_{(j)}, \ldots, x_{(p)})$
- Order node z's features: $z_o = (z_{(1)}, \ldots, z_{(j)}, \ldots, z_{(p)})$
- Construct two new nodes
  - With j: $x_{+j} = (x_{(1)}, \ldots, x_{(j-1)}, x_{(j)}, z_{(j+1)}, \ldots, z_{(p)})$
  - Without j: $x_{-j} = (x_{(1)}, \ldots, x_{(j-1)}, z_{(j)}, z_{(j+1)}, \ldots, z_{(p)})$
- Compute marginal contribution: $\phi_j^m = \hat{f}(x_{+j}) - \hat{f}(x_{-j})$

- Compute Shapley value as the average: $\phi_j(x) = \frac{1}{m}\sum_{m=1}^{M} \phi_j^m$

The resulted Shapley value can tell us whether the node feature has a negative or positive correlation with the final prediction. This is very helpful in matching generated visualization with known fraud patterns e.g. whether a low transaction amount is more likely to be associated with a fraud prediction.

### 4.5.3. Edge Missingness Shapley Values

The algorithm to approximate Shapley values for edge missingness is as follows:
- Output: Shapley value for the missingness of the $j^{th}$ edge
- Required: Number of iterations M, edge of interest x, subgraph G, and machine learning model f
  - For all m = 1,…,M:
    - Sample coalitions $z_k' \in \{0,1\}^M$ and $z_k \in \{0,1\}^M$ (1 = edge present in coalition, 0 = edge absent) where $z_k'(x) = 0$ and $z_k(x) = 1$
    - Get prediction for each coalition by first converting them to the original feature space and then re-applying model f
    - Compute marginal contribution: $\phi_j^m = \hat{f}(z_k') - \hat{f}(z_k)$
- Compute Shapley value as the average: $\phi_j = \frac{1}{m}\sum_{m=1}^{M} \phi_j^m$

The Shapley value for edge missingness can tell us if the prediction would change if the edge were removed e.g., a highly negative Shapley value for an edge means that removing the edge from the subgraph will change the prediction to the other class, and that edge contains most of the message passing contributing to the prediction of the node to explain.

# 5. Experiments and Results

## 5.1. Fraud Detection

Considering that the J.P.Morgan dataset is highly imbalanced with 1,467,358 non-fraudulent transactions and 30,819 fraudulent transactions, Average Prevision (AP) and Area under the ROC Curve (AUC) are used to eliminate potential misleading interpretation of a high accuracy result.

| Model | Accuracy | Loss | Average Precision | AUC |
|---|---|---|---|---|
| GAT | 0.8845 | 0.2234 | 0.8132 | 0.8679 |
| GEM | 0.8923 | 0.2454 | 0.8023 | 0.8865 |
| xFraud detector (original data) | 0.8946 | 0.2421 | 0.8774 | 0.8821 |
| xFraud detector without Line of Business and Sector fields | 0.9310 | 0.2312 | 0.9212 | 0.8915 |
| xFraud detector without Line of Business and Sector fields, and with *dAmount* and *dTime* fields | 0.9461 | 0.2127 | 0.9370 | 0.9640 |

Figure 15: Fraud detection results

Overall, the performance of the 3 detection models is consistent with our prediction. The 2 baseline models GAT and GEM do not perform as well as xFraud detector due to their limitations in processing type features and relations.

Based on the data analysis in Section 4.2, *dAmount* and *dTime* are expected to improve the model performance, while the original Line of Business and Sector fields may not. The results show that adding *dAmount* and *dTime* as features to transaction nodes improves the performance significantly over all 4 metrics.

### 5.2. Fraud Explanation

The proposed explainer gives us a GNNExplainer (Ying et al, 2019) subgraph, 2 bar charts for node features and edge missingness SHAP. We also added relevant original tabular data in the final visualization to provide more context for the explanation analysis.

We will analyse 2 fraud cases of 2 different transaction types (withdrawal and quick payment) and 1 non-fraud case. We expect the fraud cases will have patterns matching ATO fraud (Forbes Finance Council, 2022).

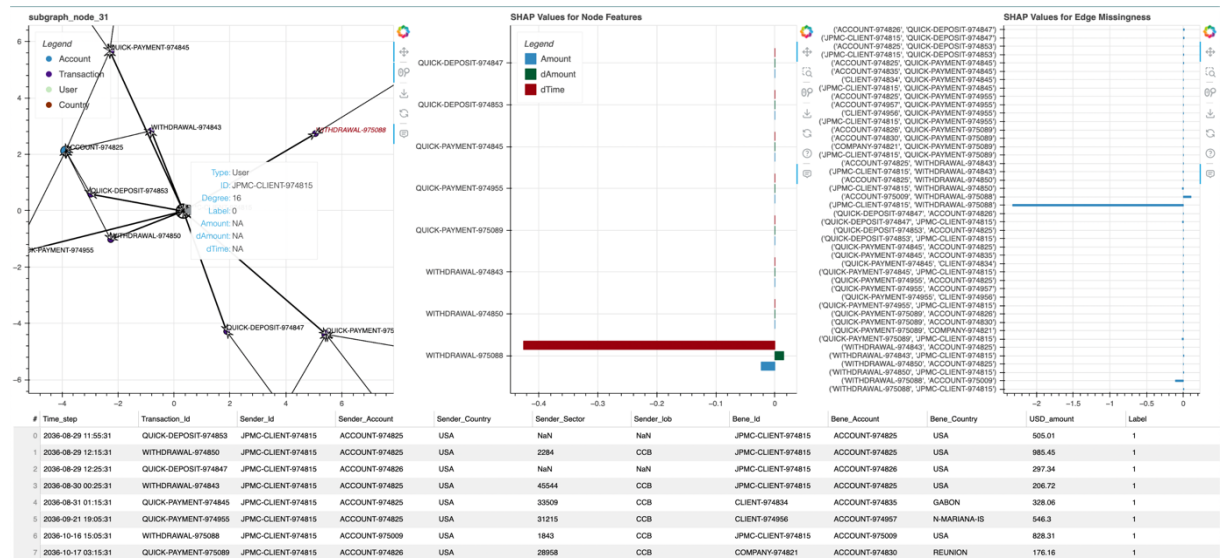

Figure 16: The visualization for the fraud case *WITHDRAWAL-975088*

In Figure 16, we can see that the edge connecting JPMC-CLIENT-974815 to the current node-to-explain WITHDRAWAL-975088 has a negative SHAP value with the highest absolute value. This means that if this edge is removed, it is more likely for the model to predict WITHDRAWAL-975088 as non-fraud.

For the node features, *dTime* has a high negative correlation with the current node's prediction. This is consistent with our analysis that fraud transactions tend to be executed rapidly, hence the lower *dTime* is, the more likely it is for the transaction to be predicted as fraud.

In the subgraph generated by GNNExplainer, we can see that JPMC-CLIENT-974815 make several successive quick payments before executing WITHDRAWAL-975088. These quick payments are sent to accounts at unusual countries like Gabon and Northern Mariana Islands. This behaviour is consistent with the ATO (Forbes Finance Council, 2021) pattern, where the fraudster quickly makes instant payments from the victim's account to another account and withdraw from there afterwards.

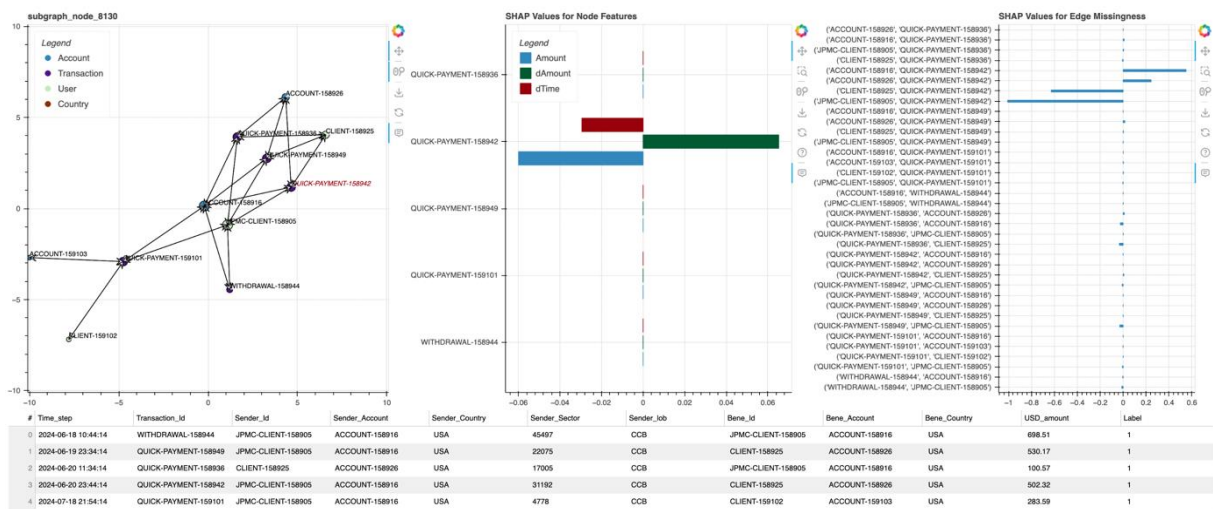

Figure 17: The visualization for the fraud case *QUICK-PAYMENT-158942*

Figure 17 shows the visualization for a fraudulent quick payment transaction. QUICK-PAYMENT-158942 appears to be a part of an ATO fraud as the client involved (JPMC-CLIENT-158905) makes several quick payments within a short period of time following a withdrawal. We can also see that CLIENT-158925 and JPMC-CLIENT-158905 send money back and forth more than once, indicating abnormal movement of money.

Looking at the node features SHAP values, *dTime* and *Amount* have negative correlation with the fraud prediction, showing that the faster transactions are executed, and the smaller the transaction amounts are, the more likely it is that involved clients are carrying out ATO fraud. In contrast, *dAmount* has a hight positive correlation with the fraud prediction, meaning that if a large transaction is carried out right after a small one, it is indicative of fraud, which is consistent with ATO whose indicators include the fraudster sending small transactions for testing purposes.

As for edge missingness SHAP values, it is similar to the case of WITHDRAWAL-975088, where most fraud indicative information comes from the clients making and receiving the relevant transactions.

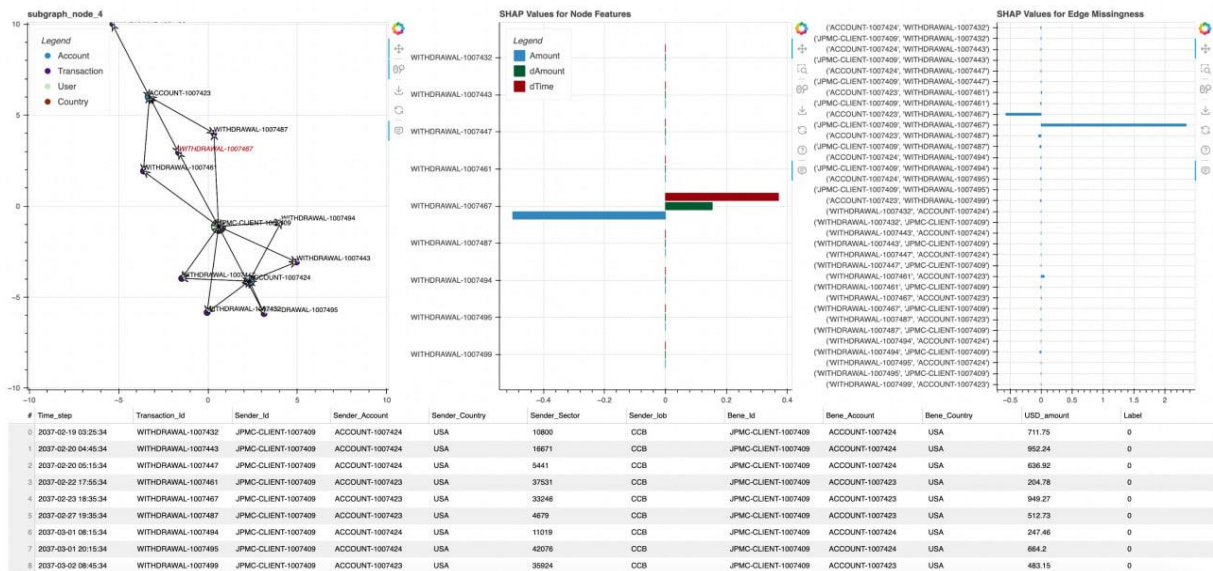

Figure 18: The visualization for the non-fraud case *WITHDRAWAL-1007467*

Figure 18 shows the visualization for a non-fraud case. The patterns of fraud cases in Figure 16 and 17 do not exist here. More specifically, JPMC-CLIENT-1007409 does not make any quick payments, and the receiver resides in USA which is a common country for a transaction.

From the edge missingness SHAP values, if the edge between JPMC-CLIENT-1007409 and WITHDRAWAL-1007467 is removed, it is unlikely for the model to change its prediction. This is reasonable since less information is passed to the transaction node, keeping its risk score low.

Regarding the node features SHAP values, only the direct features have significant impact. *dTime* has a positive correlation with the non-fraud prediction, consistent with the analysis that non-fraud accounts take more time between consecutive transactions. *Amount* has a negative correlation, showing that transactions with less USD amount tend to be non-fraudulent.

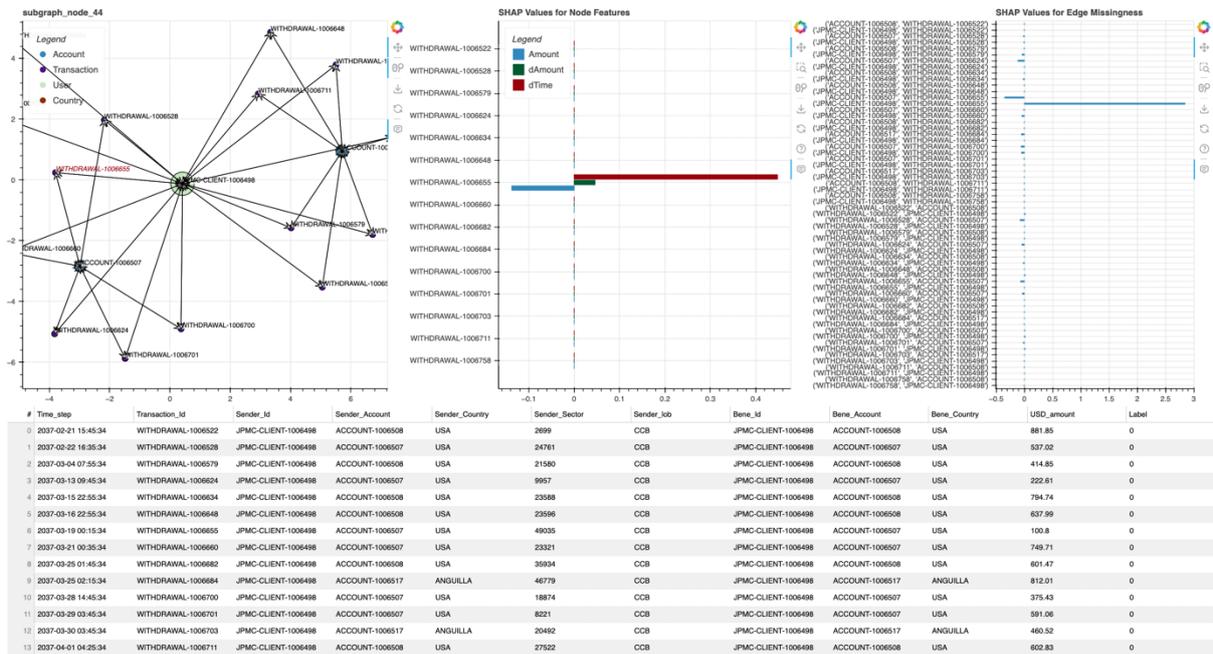

figure 19: The visualization for the non-fraud case *WITHDRAWAL-1006655*

For the case of WITHDRAWAL-1006655 in Figure 19, we can see that JPMC-CLIENT-1006498 makes transactions with 2 different accounts, 1 in USA and 1 in Anguilla. Apart from that, the rest of the visualization seems similar to the case of WITHDRAWAL-1007467 in Figure 18. This means that the detector does not think that transacting on 2 different accounts in 2 different countries is an indicator of ATO fraud. In some payment systems, this could be a decisive factor for a model to flag a transaction as fraud (Bukszpan, 2012).

## 6. Conclusion

### 6.1. Discussion

In this project, we propose a composite explainer consisting of GNNExplainer, node features and edge missingness Shapley values as a more comprehensive solution than existing explainers for GNNs.

We adapt the xFraud detector to our J.P.Morgan dataset and obtain good performance in terms of accuracy, average precision, and AUC. This performance is expected since xFraud's detection task is the same as ours. Moreover, the J.P.Morgan dataset is generated using a simulator with multi-agent system, making the fraud patterns consistent, and the interaction between agents in the simulator is fully captured with a graph dataset. We also confirm that the 2 baseline models GAT and GEM give inferior performance due to a lack of type feature processing.

Supported by knowledge of real-world fraud cases, we can analyse the generated visualization to confirm that there exists patterns indicating the flagged fraud transactions matching those from typical ATO scenarios in Section 5.2.

## 6.2. Limitation

Regarding the fraud detection task, the synthetic J.P.Morgan dataset is quite simplistic in terms of entity types and fraud patterns compared to real-world scenario where fraudulent scheme can be much more complicated, involving a lot of money movements between accounts in the host payment system as well as third-party systems; not to mention that the patterns we discover can be hidden by fraudsters e.g. instead of executing transactions rapidly, they could stretch them out in order not to stand out in front of the detector, though that could require a trade-off between safety and profits for the fraudsters.

On the explanation side, the generated visualization from the composite explainer is often only readable to data analysts or fraud investigators who possess knowledge of payment frauds. Since the visualization consists of 3 different explanation components, there is a possibility of conflicts existing among them, and it's up to the analyst or investigator to resolve such contradiction based on the actual context and their domain knowledge.

## 6.3. Future Scope

The xFraud detector does not take into temporal information from the original data since Rao et al (2020) want the model to view all transactions made by a user at any time. While the detector gives good results, incorporating temporal data into the model has the potential to improve the detector performance as well as provide another source of data for the explainer to work on. T-GNNExplainer (Xia et al, 2023) is a solution that utilizes temporal graph networks, but this method has not been tested on a real-world dataset yet, and its generated event-specific subgraphs require better visualization to be used comprehensively. Rao et al (2022) propose DyHGN utilizing temporal graph networks in fraud detection.

While this model works well on eBay's real-world datasets, explanation for DyHGN is only given by Shapley values and not GNNExplainer's explanation subgraph. Considering that the authors of DyHGN are the same authors of xFraud where GNNExplainer is used as part of the explanation, and DyHGN is the newer research without GNNExplainer being used, there must have been significant challenges preventing them from generating explanation subgraphs as for temporal graph networks.

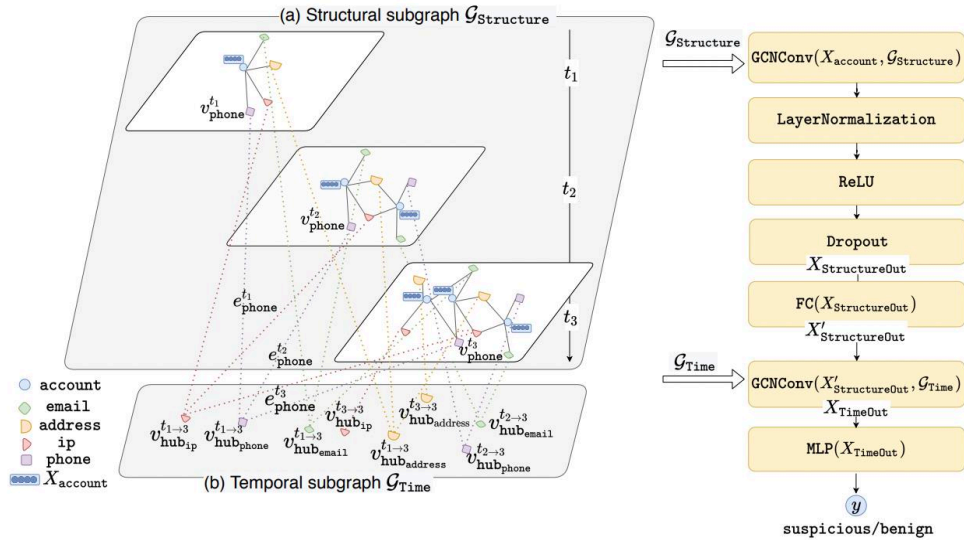

Figure 20: DyHGN architecture

This project focus on ATO fraud, but there are a myriad of fraud scenarios with varying complexity that are not captured by the synthetic J.P.Morgan dataset. Additionally, financial datasets are not easily obtained, especially datasets with semantically labelled fields as finance institutions and payment service providers usually want to privatize such information. For independent researchers, a synthetic data generator that could capture typical and modern fraud scenarios and involve more entities is desirable. PaySim (Lopez-Rojas and Axelsson, 2014) and BankSim (Lopez-Rojas, Elmir, and Axelsson, 2016) are some popular synthetic data generators for payment frauds, but their fraud scenarios are too simplistic and training a model on such data does not give convincing proof that it could be applicable in a real-world setting.